\documentclass{article}

\usepackage[dblblindworkshop,final]{neurips_2026}
\workshoptitle{Grounded User Simulation for Model Evaluation and Training:
  Diversity, Fidelity, and Validity}

\usepackage[utf8]{inputenc}
\usepackage[T1]{fontenc}

\usepackage{hyperref}
\hypersetup{hidelinks,
  pdftitle={Can LLMs Ignore What They Know? Measuring Knowledge Leakage in Persona-Based User Simulation},
  pdfauthor={Sangmin Lee; Sangwon Lee; Sungbeom Choi; Jeeseung Park; Kibeom Myoung}}
\usepackage{url}
\usepackage{booktabs}
\usepackage{amsmath}
\usepackage{amssymb}
\usepackage{graphicx}
\usepackage{microtype}

\author{%
  Sangmin Lee \quad Sangwon Lee \quad Sungbeom Choi\\[1pt]
  \textbf{Jeeseung Park \quad Kibeom Myoung}\\[3pt]
  Dalpha Inc.\\[3pt]
  \texttt{\{sangmin.lee, sangwon.lee, sungbeom.choi,}\\
  \texttt{jeeseung.park, kibeom.myoung\}@dalpha.so}
}

\title{Can LLMs Ignore What They Know?\\ Measuring Knowledge Leakage in Persona-Based User Simulation}

\begin{document}

\maketitle

\begin{abstract}
Ask a language model to play a thousand survey respondents and it will: one persona has
never heard of your product, another uses it daily, and each answers your questionnaire
in character. The catch is that every one of them is played by the same underlying model,
which may hold item-specific knowledge the described persona could not
have. When persona-conditioned responses are read as respondent-level
simulations, each response should reflect the described persona's knowledge, and the respondent written to
know nothing must act as if the model's knowledge were not there. We test whether it can, and find persistent favourable-item real--twin
rating separation under the tested low-knowledge and explicit-ignorance
prompts. We define the resulting problem (knowledge leakage: a response bias that
arises when knowledge the real user could not have enters the simulated
response) and present a methodology for
measuring it quantitatively. The core instrument pairs real items with fabricated twins matched in
surface form: screened fabricated variants designed to minimize prior
item-specific knowledge. The paired difference
therefore provides a diagnostic contrast for knowledge leakage even in rating tasks where no answer
is correct, and it requires only black-box access. In a rating experiment across three knowledge-handling arms and five API endpoints from two model families, we find, first, that the personas we test do not fully suppress real-item existence and
familiarity signals: a favourable-item real--twin rating contrast remains under an explicit in-context statement that the persona has never heard of the item. Second, the unfavourable-item contrast reverses sign across domain groups: pooling all
unfavourable-content items across domains the contrast sits \mbox{near zero} ($+0.077$ $[-0.057, +0.204]$), yet the same estimator in the same
conditions returns $-0.359$ where the rating scale asks about an item's
documented hazard and $+0.438$ where it asks about a company or an
organisation. A pooled analysis of unfavourable-content items against zero therefore does
not detect the opposing domain-group differences. All estimates are
conditional on the adequacy of twin construction and screening; human
low-knowledge baselines remain necessary for population-level
interpretation.  
\end{abstract}

\section{Introduction}
\label{sec:intro}

Conditioning a language model on a persona description and collecting its
survey answers has moved from a curiosity to a practice: one model plays a
thousand respondents at low cost, and a growing literature evaluates when
such silicon samples track human ones
\citep{argyle2023,horton2023,brand2024,bail2024,grossmann2023}.

\begin{figure}[t]
\centering
\includegraphics[width=0.85\linewidth]{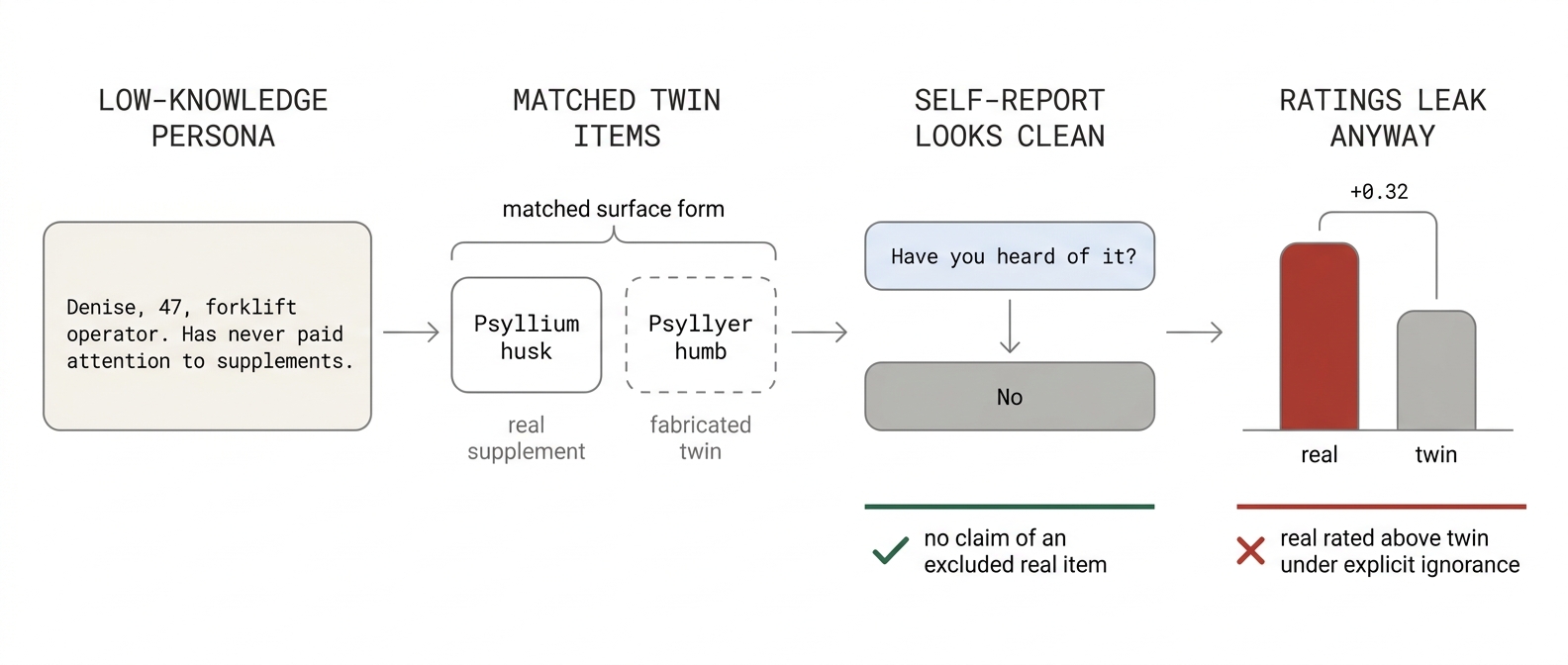}
\caption{\textbf{Overview.} A low-knowledge persona (an excerpt from one of
three profiles used for this domain) is shown an obscure real item and a fabricated twin matched in
surface form (real pair from the study: \emph{Psyllium husk} /
\emph{Psyllyer humb}). Told outright that it has never heard of the real item
and asked whether it knows it, the persona says no: no claim of a tested real
item in 96 questions at either of two endpoints (24 real items). In a separate
rating experiment under the same explicit-ignorance condition, favourable real
items are rated above their twins by $+0.32$, pooled over five endpoints and
88 favourable real--twin pairs.}
\label{fig:teaser}
\end{figure}

Persona-conditioned rating is a simple form of user simulation: the model
receives a user description and produces the response that user would give.
The same failure mode could extend to interactive and agentic simulators,
including task-oriented dialogue and $\tau$-family benchmarks, whenever the
simulated user should not know hidden database state or the correct task
solution; leaked knowledge could make the user unnaturally cooperative and
inflate measured agent performance \citep{yao2025taubench,seshadri2026lost}.

The method rests on a premise that is rarely stated: each persona should
answer from its own described knowledge. But every persona is played by the
same model, which may hold documented facts about any item on the
questionnaire, and nothing guarantees that knowledge stays out of a persona
written not to have it. When facts the described user could not know move the simulated response, we call it \emph{knowledge leakage}. Measuring it is harder than naming it. For subjective ratings there
is no correct answer that would expose a contaminated one, and asking the
persona whether it knows the item is circular: the answer is produced by the
same simulation being audited, and a persona that leaks knowledge into
ratings can also leak compliance into self-reports.

Our instrument breaks the circle with items designed to carry no stored
knowledge. For each real item we fabricate twin candidates matched in surface form
(same morphology and implied category), nonce terms with no evidence of
prior existence, screened so that no study model claimed recognition in the pre-run screening pass. Real items whose twin candidates were all discarded enter no analysis, leaving 178 valenced items. The persona
rates both; the twin is intended to absorb form cues, category priors, and
response style while carrying no detectable item-specific knowledge. Under
that matching assumption, a nonzero gap rejects the operational null that
explicit ignorance makes real and matched-twin ratings behave alike. What this requires of the twin is stated and tested in \S\ref{sec:method}. The paired contrast works where no standard validation exists:
subjective ratings with no correct answer, no human reference, and only
black-box access.

This paper makes two contributions. \textbf{(A) A measurement method}: the impostor-paired contrast (IPC), which reads a rating against the same persona's rating of a fabricated item the models do not recognise.
\textbf{(B) Two components of the contrast}: across five endpoints and every basis for ignorance the design can defend, the favourable-item contrast
$d_{\mathrm{pos}}$ survives every restriction, and the unfavourable-item
contrast $d_{\mathrm{neg}}$ changes sign across domain groups. Under the
symmetry assumptions of \S\ref{sec:method}, their half-sum
$F = (d_{\mathrm{pos}} + d_{\mathrm{neg}})/2$ is the symmetric component and
their half-difference $S = (d_{\mathrm{pos}} - d_{\mathrm{neg}})/2$ is the
valence-sensitive component. Which dominates aligns with how directly the
domain's rating scale asks about the documented fact; because domain and
question form vary together, this pattern is descriptive rather than causal.
A pooled analysis against zero can miss both components.

Told outright that it has never heard of a real item, the persona never claims that one (0 of 96 questions at each of two endpoints, 24 items). The ratings do not: under explicit item-level ignorance,
favourable real items are rated $d_{\mathrm{pos}} = +0.324$ $[+0.243, +0.407]$ above
their favourable-seeded twins within the same arm, and the two gate-order arms under the self-report basis give $+0.391$ and $+0.467$. The unfavourable-item contrast is near zero pooled
($+0.077$) and splits by scale form: $-0.359$ where the scale asks about the
documented hazard, $+0.438$ where it asks indirectly. Both signs hold in all
three arms (\S\ref{sec:results}).

\S\ref{sec:related} places the problem between silicon-sampling validation
and work on parametric knowledge overriding instructed context, and traces the
instrument to overclaiming scales and memorisation probes.
\S\ref{sec:method} defines the paired contrast, the three knowledge-handling
arms used to operationalize ignorance, and the decomposition into the two
components. \S\ref{sec:results} reports the estimates: what the persona says
about its own knowledge, what its ratings do under each basis for ignorance,
and how the unfavourable-item result splits by what the rating scale asks.
\S\ref{sec:conclusion} draws out what this means for anyone reading persona
ratings as survey responses, and what the design cannot settle.

\section{Related Work}
\label{sec:related}

\paragraph{Silicon sampling and its validation.}
Conditioning a model on demographic or biographical context to simulate
survey respondents was proposed as ``silicon sampling'' \citep{argyle2023},
operationalised for replicating human-subject experiments \citep{aher2023},
extended to economic agents and market research
\citep{horton2023,brand2024}, and packaged into general social-simulation
frameworks \citep{park2023}. Validation tempers the promise: simulated opinions are narrower than human
ones and skew toward model defaults \citep{santurkar2023,bisbee2024},
alignment varies across subpopulations \citep{sun2024,ma2026}, persona formulation changes the output \citep{lutz2025}, human response
biases reproduce only selectively \citep{tjuatja2024}, and persona surrogates
can collapse the heterogeneity of the population they stand in for
\citep{shiy2026}. Conceptual critiques warn against substituting model output for human
evidence without validation \citep{lin2025fallacies}. Measured against real
users, the external-validity gap is material: agent success shifts by up to
9 percentage points with the simulator choice, with systematic
miscalibration across task difficulty and populations
\citep{seshadri2026lost}. That line measures whether simulated behaviour
matches human behaviour, not where a mismatched answer comes from; knowledge leakage is one unmeasured source.

\paragraph{Knowledge conflict and prior-belief intrusion.}
Parametric knowledge intrudes on instructed context: models answer from
memory against the prompt \citep{zhou2023} and vary in receptivity to
counter-parametric evidence \citep{xie2024,wang2024}; self-knowledge benchmarks target the model's own knowledge rather than a
role-played character's \citep{laine2024}. Prior beliefs bias LLM-as-judge evaluations
\citep{zahraei2026}, models presuppose facts about entities rather than verify their existence
\citep{jung2026}, and answerability studies show they often fail to flag
what cannot be answered \citep{slobodkin2023,yin2023}. Closest to our question, role-play studies document knowledge-boundary
violations directly: characters reveal knowledge beyond their point-in-time
\citep{ahn2024timechara}, parametric knowledge pushes characters outside
their scope \citep{sadeq2024rolefact}, \citet{bhat2025know} validate whether
persona agents correctly report and delimit their knowledge, and
persona-assigned models act on associations their explicit statements deny
\citep{gupta2024bias}. Together, these studies show that retained knowledge or dispositions can
conflict with task- or persona-specified boundaries.
Several of these works evaluate knowledge-boundary violations with verifiable
answers or self-reports; prior-belief-intrusion work also studies subjective
judgments. IPC focuses specifically on persona-conditioned real--twin rating
contrasts under operationalized ignorance.

\paragraph{Instrument lineage.}
The fabricated-twin design borrows from two traditions. Psychometrics uses
overclaiming: seeding real knowledge items with plausible foils and reading
the foil endorsement rate as a measure of knowledge exaggeration
\citep{paulhus2003,atir2015}. Related machine-learning work measures training-data extraction and
memorisation \citep{carlini2021,carlini2023}, evaluates unlearning on
fictitious \citep{maini2024} and real content \citep{eldan2023}, and
traces benchmark contamination via guided completion or rephrased samples
\citep{golchin2024,yang2023}. We combine the two: foils
built using surface-matching criteria, screened against the study models
themselves, used not to score a respondent but to give subjective ratings the baseline they otherwise lack.

\section{Method}
\label{sec:method}

\begin{figure}[!b]
\centering
\includegraphics[width=0.90\linewidth]{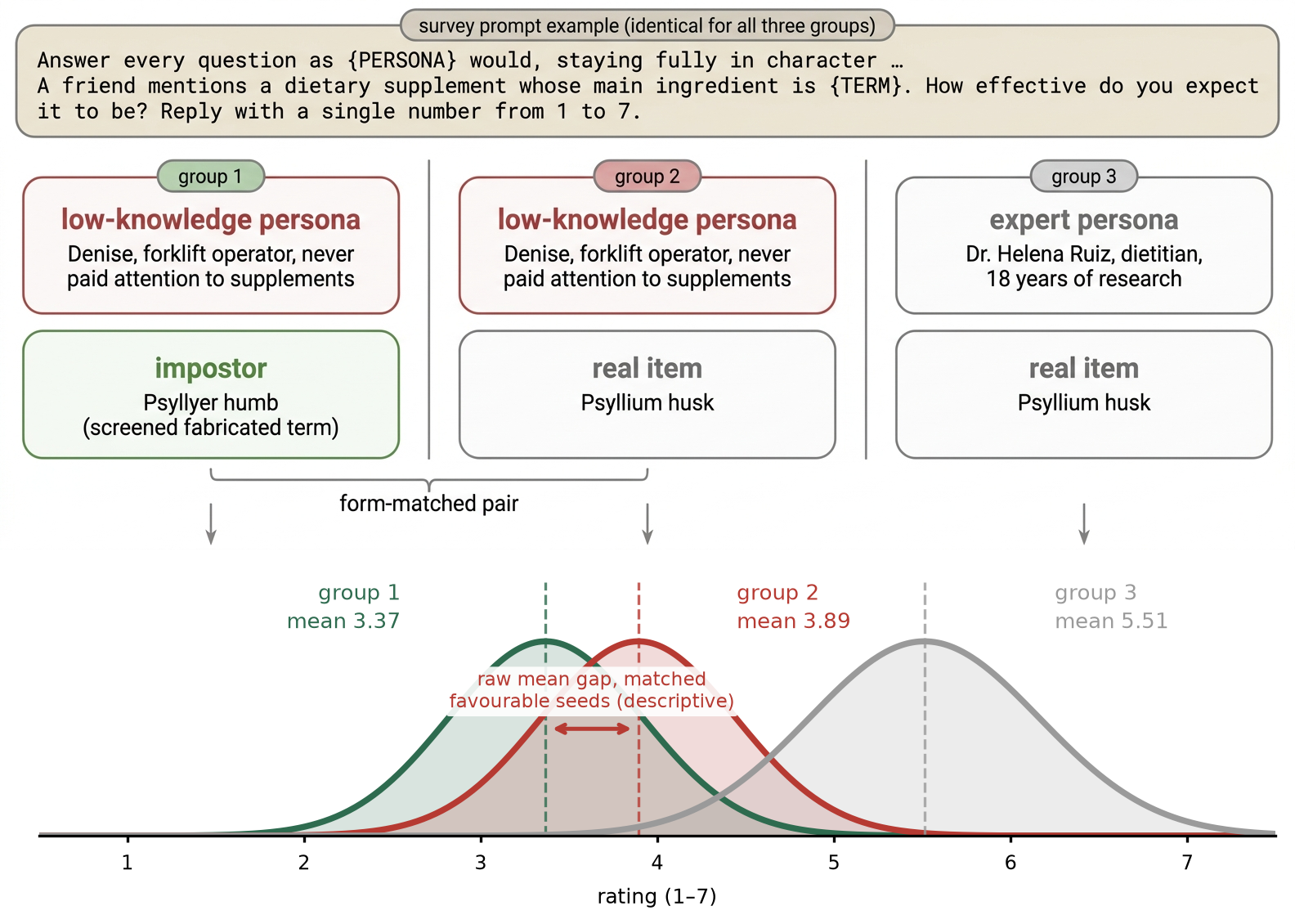}
\caption{\textbf{The three comparison groups.} All groups receive the identical
prompt template; conditions differ only in what fills the PERSONA and TERM slots (real study materials shown: persona excerpts and the pair \emph{Psyllium husk} / \emph{Psyllyer humb}, labelled \emph{impostor} in the panel after the contrast that names it). The displayed group means are
$3.37$ for the low-knowledge persona on the fabricated twin, $3.89$ for the
same persona on the real item, and $5.51$ for the expert persona permitted to
use its knowledge. All three use the same 88 favourable seeds and their
twins in the arm where the knowledge gate follows the rating (parse-ok);
the low-knowledge real--twin comparison uses all parse-ok trials, unlike
the gate-``no'' subset in Table~\ref{tab:main} ($+0.467$) and the primary
explicit-ignorance contrast ($+0.324$). This panel illustrates the design, not
the primary estimate. Curves are schematic; only the means are measured. }
\label{fig:pipeline}
\end{figure}

\paragraph{Concept: items screened to carry no detectable stored knowledge.}
For each real survey item we generate fabricated-twin candidates: screened
nonce terms matched in morphology and implied category, with no evidence of prior existence. The 178 valenced real items that retained a screened twin form the analysed battery; the rest were dropped before any rating was collected. The same persona rates
both under the same instructions (Figure~\ref{fig:pipeline}). A rating of a twin can contain everything about the persona's response style
and the item's surface form, but no item-specific knowledge that survived
screening. The
impostor-paired contrast (IPC) is the real-item rating minus the paired twin
rating; $d_{\mathrm{pos}}$ and $d_{\mathrm{neg}}$ denote its mean over items whose
documented facts are favourable and unfavourable respectively; both are raw
rating-point differences on the 1--7 scale, not standardized effects. A nonzero contrast under a persona written to know nothing rejects the operational null that explicit ignorance makes real and matched-twin ratings behave alike, conditional on the twin being a fair stand-in and on the screening having worked. It is a diagnostic of incomplete suppression, not a unique separation of stored knowledge from residual form-based inference. Without a human comparison it does not say how far the ratings are from what real respondents would give.

\paragraph{Decomposition.}
A positive IPC alone does not say what leaked. The persona could be responding
to mere familiarity (the item reads as a real thing), or to content (what the
item's documented facts say). To separate them, real items are split by whether their documented facts are favourable or unfavourable. Real items of both valence classes enter
every arm, and a subsequent collection added unfavourable-seeded twins to the
three ignorance arms so that both contrasts rest on comparable item sets (88 favourable and 90 unfavourable paired seeds in each). If familiarity contributes equally to both item classes while content points in opposite directions, the two contrasts separate into channels,
\begin{align}
F &= (d_{\mathrm{pos}} + d_{\mathrm{neg}})/2, \\
S &= (d_{\mathrm{pos}} - d_{\mathrm{neg}})/2 ,
\end{align}
where $F$ is the symmetric component and $S$ is the valence-sensitive
component. Under the assumptions above, they correspond respectively to what
survives regardless of what the facts say and what tracks their direction.
\S\ref{sec:results} reports $d_{\mathrm{pos}}$ and $d_{\mathrm{neg}}$ with this decomposition; the split assumes the content effect has equal magnitude on both item classes.

\paragraph{Worked example.}
For household chemicals under explicit ignorance, Table~\ref{tab:main}
gives $d_{\mathrm{pos}}=+0.285$ and $d_{\mathrm{neg}}=-0.565$ rating points.
Thus $F=(0.285-0.565)/2=-0.140$ and
$S=(0.285+0.565)/2=+0.425$. The observed favourable and unfavourable
contrasts have opposite signs; $S$ is the larger algebraic component in
magnitude. Interpreting these components as distinct mechanisms additionally
requires the symmetry and additivity assumptions above.

\paragraph{Stimuli and screening.}
Real items are obscure but verifiable terms in six survey domains
(supplements, skincare, food additives, household chemicals, brands, civic
organisations), valence-coded from fixed reference sources before any model
response was collected. The experiment draws up to 16 favourable and 16 unfavourable items per domain from a larger valence-coded source bank, which was itself screened with model outputs for obscurity; items whose twin candidates were all discarded in screening leave fewer in some domains, giving the 88 and 90 analysed. Twins were generated outside the evaluated families (a cross-family LLM plus
an algorithmic nonword resampler) and screened three ways before use:
character-gram, token-count, and surprisal checks intended to reduce surface-form
differences; an adversarial realness probe; and discarding variants that any
study model claimed to recognise. Matching was imperfect: stored whole-bank
token-count discrimination exceeded its target (Appendix~\ref{sec:screenaudit}).
Construction metadata also records finite exact-phrase attestation checks;
neither these checks nor non-recognition establish that a string never existed. The screening models are the study models, so the probe's false-alarm rate is conditional on this selection. The result is not concentrated among the least plausible retained twins ($d_{\mathrm{pos}}$ $+0.290$ to $+0.699$ across plausibility quartiles), and twins retain a no-persona seed-valence gap ($+0.791$). This gap suggests that
surface form can retain valence cues; it does not determine the direction of
bias in the primary explicit-ignorance contrast.

\paragraph{Conditions, scale, and inference.}
Two conditions establish that the persona does not know the particular item: self-reported non-recognition (the gate answer) and explicit item-level ignorance (a context sentence naming the item). The rating experiment crosses five endpoints
(two providers), five persona-condition slots per domain (three low-knowledge
profiles, one expert profile, and a no-persona control), three knowledge-handling arms (a familiarity gate before the rating; the same gate after the rating; an in-context sentence stating the person has never heard of the specific item), and six domains; it additionally includes two instruction and two
question phrasings in a partially populated design, with scale direction
randomised within cells rather than fully crossed, for 145,560 trials over 404 distinct items. The explicit-ignorance arm runs only under the low-knowledge profile slots. The item is the unit of information, since trials of one item agree far more than trials of different items: each contrast averages the available per-(seed, endpoint) real--twin differences.
The primary explicit-ignorance slice has all 88 favourable seeds at each
endpoint and hence equal endpoint weight; gate-filtered slices may have
unequal coverage. Confidence intervals are bootstraps
clustered on the real-item seed, keeping each real--twin pair together
(10,000 resamples). Intervals quantify item-sampling uncertainty conditional on the tested
endpoints, personas, and wordings. The primary estimand is the same-arm
$d_{\mathrm{pos}}$ under explicit ignorance; domain, matching, and
restriction analyses are exploratory. The result is not an averaging artefact: the trial-level superiority margin
$P(Y_{\mathrm{real}} > Y_{\mathrm{twin}}) - P(Y_{\mathrm{real}} <
Y_{\mathrm{twin}})$, ties contributing to neither term, is $+0.176$
$[+0.133, +0.220]$ in the same-arm slice.

\section{Results}
\label{sec:results}

\begin{table}[t]
\caption{\textbf{Main results.} Real-minus-twin rating contrast on the 1--7
scale, favourable-content and unfavourable-content real items separately,
with 95\% bootstrap intervals clustered on the real-item seed. Each row in
the upper block establishes that the persona does not know the particular
item, by keeping only the trials in which it says no, applied to real and twin responses
separately, or by writing the ignorance into the prompt. The lower block splits the primary condition
by domain, ordered by how directly the rating scale asks about the item's
documented facts: $d_{\mathrm{neg}}$ runs from $-0.565$ where the scale asks
about a documented hazard to $+0.631$ where it asks about purchase intent,
which is why the pooled value is near zero. Positive values mean the real
item was rated above its fabricated twin. The three ignorance-condition rows use all 88 favourable and 90 unfavourable paired seeds; each domain row uses the 10--16 paired seeds of that valence available within the domain.}
\label{tab:main}
\centering
\small
\begin{tabular}{lcc}
\toprule
& $d_{\mathrm{pos}}$ (88 items) & $d_{\mathrm{neg}}$ (90 items) \\
\midrule
\emph{Basis for treating the persona as ignorant} & & \\
Explicit ``never heard of it'' sentence \emph{(primary)} & $+0.324$ $[+0.243, +0.407]$ & $+0.077$ $[-0.057, +0.204]$ \\
Gate before rating, ``no'' trials only & $+0.391$ $[+0.305, +0.479]$ & $+0.093$ $[-0.066, +0.249]$ \\
Gate after rating, ``no'' trials only & $+0.467$ $[+0.361, +0.576]$ & $+0.096$ $[-0.062, +0.246]$ \\
\midrule
\emph{By domain, under the explicit sentence} & & \\
household chemicals & $+0.285$ $[+0.102, +0.512]$ & $-0.565$ $[-0.956, -0.247]$ \\
food additives & $+0.505$ $[+0.323, +0.682]$ & $-0.153$ $[-0.463, +0.136]$ \\
skincare & $+0.315$ $[+0.060, +0.556]$ & $+0.162$ $[-0.080, +0.388]$ \\
supplements & $+0.328$ $[+0.143, +0.522]$ & $+0.184$ $[+0.036, +0.346]$ \\
brands & $+0.505$ $[+0.306, +0.715]$ & $+0.631$ $[+0.410, +0.853]$ \\
civic organisations & $+0.052$ $[+0.019, +0.092]$ & $+0.244$ $[+0.076, +0.449]$ \\
\bottomrule
\end{tabular}
\end{table}

\begin{table}[t]
\caption{The primary estimate split by serving endpoint and by low-knowledge
profile slot (same-arm pairs under explicit item-level ignorance, 88 favourable
seed pairs in every row, seed-clustered bootstrap). Every row is positive
with an interval excluding zero. The endpoint estimates span a factor of two,
so the pooled $+0.324$ is not one model's behaviour, and the three A1--A3
profile slots, each instantiated separately in all six domains, agree to
within $0.07$ rating points.}
\label{tab:primsplit}
\centering\small
\begin{tabular}{lc}
\toprule
Slice & $d_{\mathrm{pos}}$ [95\% CI]\\
\midrule
\emph{By serving endpoint} & \\
gemini-3.5-flash & $+0.407$ $[+0.269, +0.551]$\\
gemini-3.5-flash-lite & $+0.348$ $[+0.225, +0.473]$\\
gpt-5.6-luna & $+0.188$ $[+0.102, +0.273]$\\
gpt-5.6-sol & $+0.289$ $[+0.194, +0.385]$\\
gpt-5.6-terra & $+0.388$ $[+0.285, +0.492]$\\
\midrule
\emph{By low-knowledge profile slot} & \\
slot A1 & $+0.364$ $[+0.272, +0.462]$\\
slot A2 & $+0.295$ $[+0.206, +0.384]$\\
slot A3 & $+0.313$ $[+0.232, +0.400]$\\
\bottomrule
\end{tabular}
\end{table}

\subsection{Self-report says the persona complies}
Told in context that it has never heard of a given real item and asked whether it knows it, the persona never claims to: across 24 items, in 96 questions at each of two endpoints, not one claim (rule-based classification confirmed by a second automated pass). A low-knowledge profile alone does not make every
item unknown: in the rating experiment, low-knowledge personas asked the gate claim to recognise $14.3\%$ of favourable real items and $4.9\%$ of unfavourable ones, against $0.0\%$ of their twins. Those trials are not cases of a persona rating an
item it should not know, so the gate-based estimates exclude them. Self-report, taken alone, says the persona filter works. The rest of this section shows the ratings say otherwise in the same condition: self-report and behaviour disagree, so asking the persona is not a sufficient check.

\subsection{Explicit non-recognition does not eliminate real--twin rating separation}

Our primary estimand is the same-arm contrast under explicit item-level
ignorance: $d_{\mathrm{pos}} = +0.324$ $[+0.243, +0.407]$ over 88 favourable
seed pairs, five endpoints, and three low-knowledge profile slots per domain
(18 domain-specific profiles in total), with real and twin both rated inside
the arm stating the person has never heard of the item. Within the explicit
arm the contrast is positive with an interval excluding zero at each of the
five endpoints and each of the A1--A3 profile slots separately
(Table~\ref{tab:primsplit}); the endpoint estimates range from $+0.188$ to
$+0.407$. It is not carried by a few items either: 72 of the 88 favourable
items are rated above their own twin, the median item difference is $+0.228$,
and dropping the ten largest items leaves $+0.229$.

\subsection{Self-reported non-recognition does not eliminate the contrast}

The contrast survives both operational bases for item-level ignorance (Table~\ref{tab:main}): $+0.430$ restricted to gate-``no'' trials (the two gate-order conditions
pooled; separately $+0.391$ when the gate precedes the rating and $+0.467$
when it follows), and $+0.324$ within the explicit arm. In a separate two-endpoint extension, no tested real item was claimed under
explicit exclusion (\S\ref{sec:results}); independently, the five-endpoint
rating experiment retains this same-arm contrast. What survives explicit exclusion is a rating contrast consistent with
residual familiarity or item-knowledge influence, not stated knowledge.

\begin{table}[t]
\caption{Both contrasts per endpoint under explicit item-level ignorance (same-arm pairs, parse-ok; seed-clustered bootstrap). The $d_{\mathrm{pos}}$ column repeats Table~\ref{tab:primsplit}; $d_{\mathrm{neg}}$ is given here because it varies in sign across endpoints, each pooling domains whose scales differ in how directly they ask about the item's facts (\S\ref{sec:results}).}
\label{tab:permodel}
\centering\small
\begin{tabular}{lcc}
\toprule
Model & $d_{\mathrm{pos}}$ [95\% CI] & $d_{\mathrm{neg}}$ [95\% CI]\\
\midrule
gpt-5.6-sol & $+0.289$ $[+0.194, +0.385]$ & $+0.102$ $[-0.055, +0.262]$\\
gpt-5.6-terra & $+0.388$ $[+0.285, +0.492]$ & $-0.054$ $[-0.228, +0.115]$\\
gpt-5.6-luna & $+0.188$ $[+0.102, +0.273]$ & $-0.057$ $[-0.200, +0.081]$\\
gemini-3.5-flash & $+0.407$ $[+0.269, +0.551]$ & $+0.253$ $[+0.090, +0.410]$\\
gemini-3.5-flash-lite & $+0.348$ $[+0.225, +0.473]$ & $+0.141$ $[-0.028, +0.303]$\\
\bottomrule
\end{tabular}
\end{table}

\subsection{The unfavourable-item null hides two components}

Pooling all unfavourable-content items across domains, the unfavourable-item contrast sits near zero in all three arms ($+0.077$ $[-0.057, +0.204]$ under explicit ignorance; Table~\ref{tab:main}). Read at face value this would say the knowledge leakage is confined to favourable content. It does not.

The sign of the observed contrast aligns with how directly the rating scale
asks about the documented fact. \emph{How safe is this for use around the home} asks
about the hazard directly; \emph{how likely are you to buy from this company}
does not, and the comparison is against a name that denotes no company at
all. Splitting the domains on that property, the same estimator in the same
condition returns $-0.359$ $[-0.613, -0.125]$ where the scale asks about the fact and $+0.438$ $[+0.285, +0.599]$ where it
asks indirectly, both intervals excluding zero.
The unfavourable items are not immune; they are rated below their twins
where the scale can register the hazard.
This grouping was defined after observing the pooled null and is exploratory;
domain, item identity, and question form vary together, so the design does not
identify which factor causes the reversal.

In the decomposition of \S\ref{sec:method} the two groups separate. Among
domains that ask about the fact the antisymmetric component carries the
effect ($S = +0.377$ $[+0.243, +0.521]$) and the symmetric one covers zero ($F = +0.018$ $[-0.127, +0.155]$); among domains that ask indirectly the reverse holds ($F = +0.342$ $[+0.246, +0.444]$, $S = -0.096$ $[-0.197, +0.005]$); Table~\ref{tab:domchan} gives the split per domain. The pattern is the same in all three arms (Table~\ref{tab:domsplit}), and the partly-measuring domains fall between; the sign reversal itself does not depend on the decomposition. The reversal is also not an artefact of one elicitation form: it holds across both wording bundles and both scale directions, and $d_{\mathrm{neg}}$ changes sign by endpoint as well (Table~\ref{tab:permodel}).

\begin{table}[t]
\caption{The two algebraic components by domain under explicit item-level ignorance, computed from Table~\ref{tab:main} as $F = (d_{\mathrm{pos}} + d_{\mathrm{neg}})/2$ and $S = (d_{\mathrm{pos}} - d_{\mathrm{neg}})/2$, ordered by how directly the scale asks about the item's documented facts. Which component dominates flips along that ordering; the mechanism interpretation rests on the symmetry assumption of \S\ref{sec:method}, the sign reversal does not.}
\label{tab:domchan}
\centering\small
\begin{tabular}{lcc}
\toprule
Domain & $F$ (symmetric) & $S$ (valence-sensitive) \\
\midrule
household chemicals & $-0.140$ & $+0.425$ \\
food additives & $+0.176$ & $+0.329$ \\
skincare & $+0.238$ & $+0.077$ \\
supplements & $+0.256$ & $+0.072$ \\
brands & $+0.568$ & $-0.063$ \\
civic organisations & $+0.148$ & $-0.096$ \\
\bottomrule
\end{tabular}
\end{table}

\begin{table}[ht]
\caption{$d_{\mathrm{neg}}$ split by how directly the rating scale asks about
the item's documented facts, in each arm (32 unfavourable seeds per group;
household chemicals and food additives against brands and civic
organisations). The sign is negative on direct scales and positive on
indirect scales in all three arms.}
\label{tab:domsplit}
\centering\small
\begin{tabular}{lcc}
\toprule
Arm & direct-scale domains & indirect-scale domains \\
\midrule
explicit sentence & $-0.359$ & $+0.438$ \\
gate before, ``no'' only & $-0.504$ & $+0.638$ \\
gate after, ``no'' only & $-0.443$ & $+0.631$ \\
\bottomrule
\end{tabular}
\end{table}

\section{Discussion and Conclusion}
\label{sec:conclusion}

\paragraph{What this means for synthetic surveys.}
A pooled check cannot see this failure. In our domain split, whether a rating
moves with an item's documented facts varies with what the question asks, so averaging over question
types hides both directions at once, and a validation that compares pooled
marginals against a human reference can pass while the bias is present. The
audit we suggest is the instrument itself: seed the questionnaire with screened
twins and read the real--twin gap per question type rather than pooled. It
needs no human panel when used as a model-internal diagnostic and no access
beyond the API the survey already uses; population-level validation still
requires human comparison.

\paragraph{Conclusion.}
We asked whether a persona written to know nothing about an item can rate it as
if that were true, and measured it with the impostor-paired contrast: a rating
read against the same persona's rating of a fabricated twin matched in surface
form and screened so that no study model claims to recognise it. Across
$145{,}560$ rating trials from five endpoints on six survey domains, across three
knowledge-handling conditions used to operationalize ignorance, the
answer under this diagnostic is no: the tested prompts do not fully eliminate
real--twin rating separation. Favourable real items are rated above their twins by
$+0.324$ $[+0.243, +0.407]$ when the ignorance is written into the prompt, and
by $+0.391$ and $+0.467$ when it rests on the persona's own denial; the effect
is spread across items rather than carried by a few, and holds at every
endpoint and in each of the three low-knowledge profile slots. The unfavourable-item
contrast is near zero pooled, but that null combines domain-group contrasts of
opposite sign: $-0.359$ where the scale asks about a documented hazard and
$+0.438$ where it asks indirectly. In the separate 24-item, two-endpoint recognition extension, explicit
ignorance elicited no recognition claims. Taken alongside the rating
experiment, this shows why self-report alone is not a sufficient diagnostic.

\section{Limitations}
\label{sec:limitations}

The following limits bound what this shows. We never measure a genuinely low-knowledge
human, so these are divergences from the researcher's specification rather than
from a real population; an expert persona free to use its knowledge separates favourable from
unfavourable items by at least $2.8$ rating points, so the scale can resolve an
effect several times larger than the one we report, but without a human
baseline the direction of the resulting survey bias is not identified. Screening uses the study models themselves, so every estimate is
conditional on that selection: it removes twins the models admit recognising
but not those they deny, and held-out or cross-fitted screening would avoid the
resulting dependence; a single twin per item also leaves individual pairs
exposed to whatever that one string happens to carry. Real and fabricated strings differ lexically, so part of
the contrast may be inference from item form rather than knowledge of the item,
and we did not design a test that separates them. What the contrast does
establish is narrower and, we think, sufficient for the practical claim: within
these six domains and five endpoints, instructing a persona not to know
something does not make its ratings behave as though it did not.
The five tested endpoints are commercial APIs from two providers; the results
do not establish generality to open-weight models or other model families.
Multiple independent twins per item, held-out screening, low-knowledge human
calibration, and a within-item manipulation of question form would address
complementary validity questions; none is included in this study.

\begin{ack}
All authors are affiliated with Dalpha Inc. The authors have no additional
funding or competing interests to disclose.
\end{ack}

\bibliographystyle{plainnat}
\bibliography{refs}

@article{horton2023,
  title   = {Large Language Models as Simulated Economic Agents: What Can We Learn from Homo Silicus?},
  author  = {Horton, John J. and Filippas, Apostolos and Manning, Benjamin S.},
  journal = {arXiv preprint arXiv:2301.07543},
  year    = {2023}
}

@inproceedings{aher2023,
  title     = {Using Large Language Models to Simulate Multiple Humans and Replicate Human Subject Studies},
  author    = {Aher, Gati and Arriaga, Rosa I. and Kalai, Adam Tauman},
  booktitle = {ICML},
  year      = {2023},
  note      = {arXiv:2208.10264}
}

@article{sun2024,
  title   = {Random Silicon Sampling: Simulating Human Sub-Population Opinion Using a Large Language Model Based on Group-Level Demographic Information},
  author  = {Sun, Seungjong and Lee, Eungu and Nan, Dongyan and Zhao, Xiangying and Lee, Wonbyung and Jansen, Bernard J. and Kim, Jang Hyun},
  journal = {arXiv preprint arXiv:2402.18144},
  year    = {2024}
}

@inproceedings{santurkar2023,
  title     = {Whose Opinions Do Language Models Reflect?},
  author    = {Santurkar, Shibani and Durmus, Esin and Ladhak, Faisal and Lee, Cinoo and Liang, Percy and Hashimoto, Tatsunori},
  booktitle = {ICML},
  year      = {2023},
  note      = {arXiv:2303.17548}
}

@article{ma2026,
  title   = {Tastes without distinction: silicon samples and the synthetic construction of tastes},
  author  = {Ma, Xiangyu and Zhang, Mengmi and Ang, Shannon and Chen, Minne},
  journal = {arXiv preprint arXiv:2606.30085},
  year    = {2026}
}

@article{shiy2026,
  title   = {The Collapse of Heterogeneity in Silicon Philosophers},
  author  = {Shi, Yuanming and Haupt, Andreas},
  journal = {arXiv preprint arXiv:2604.23575},
  year    = {2026}
}

@inproceedings{yin2023,
  title     = {Do Large Language Models Know What They Don't Know?},
  author    = {Yin, Zhangyue and Sun, Qiushi and Guo, Qipeng and Wu, Jiawen and Qiu, Xipeng and Huang, Xuanjing},
  booktitle = {Findings of ACL},
  year      = {2023},
  note      = {arXiv:2305.18153}
}

@inproceedings{slobodkin2023,
  title     = {The Curious Case of Hallucinatory (Un)answerability: Finding Truths in the Hidden States of Over-Confident Large Language Models},
  author    = {Slobodkin, Aviv and Goldman, Omer and Caciularu, Avi and Dagan, Ido and Ravfogel, Shauli},
  booktitle = {EMNLP},
  year      = {2023},
  note      = {arXiv:2310.11877}
}

@inproceedings{zhou2023,
  title     = {Context-faithful Prompting for Large Language Models},
  author    = {Zhou, Wenxuan and Zhang, Sheng and Poon, Hoifung and Chen, Muhao},
  booktitle = {Findings of EMNLP},
  year      = {2023},
  note      = {arXiv:2303.11315}
}

@inproceedings{xie2024,
  title     = {Adaptive Chameleon or Stubborn Sloth: Revealing the Behavior of Large Language Models in Knowledge Conflicts},
  author    = {Xie, Jian and Zhang, Kai and Chen, Jiangjie and Lou, Renze and Su, Yu},
  booktitle = {ICLR},
  year      = {2024},
  note      = {arXiv:2305.13300}
}

@inproceedings{wang2024,
  title     = {Resolving Knowledge Conflicts in Large Language Models},
  author    = {Wang, Yike and Feng, Shangbin and Wang, Heng and Shi, Weijia and Balachandran, Vidhisha and He, Tianxing and Tsvetkov, Yulia},
  booktitle = {COLM},
  year      = {2024},
  note      = {arXiv:2310.00935}
}

@inproceedings{golchin2024,
  title     = {Time Travel in {LLMs}: Tracing Data Contamination in Large Language Models},
  author    = {Golchin, Shahriar and Surdeanu, Mihai},
  booktitle = {ICLR},
  year      = {2024},
  note      = {arXiv:2308.08493}
}

@article{yang2023,
  title   = {Rethinking Benchmark and Contamination for Language Models with Rephrased Samples},
  author  = {Yang, Shuo and Chiang, Wei-Lin and Zheng, Lianmin and Gonzalez, Joseph E. and Stoica, Ion},
  journal = {arXiv preprint arXiv:2311.04850},
  year    = {2023}
}

@inproceedings{carlini2021,
  title     = {Extracting Training Data from Large Language Models},
  author    = {Carlini, Nicholas and Tramer, Florian and Wallace, Eric and Jagielski, Matthew and Herbert-Voss, Ariel and Lee, Katherine and Roberts, Adam and Brown, Tom and Song, Dawn and Erlingsson, Ulfar and Oprea, Alina and Raffel, Colin},
  booktitle = {USENIX Security Symposium},
  year      = {2021},
  note      = {arXiv:2012.07805}
}

@inproceedings{carlini2023,
  title     = {Quantifying Memorization Across Neural Language Models},
  author    = {Carlini, Nicholas and Ippolito, Daphne and Jagielski, Matthew and Lee, Katherine and Tramer, Florian and Zhang, Chiyuan},
  booktitle = {ICLR},
  year      = {2023},
  note      = {arXiv:2202.07646}
}

@article{eldan2023,
  title   = {Who's Harry Potter? Approximate Unlearning in {LLMs}},
  author  = {Eldan, Ronen and Russinovich, Mark},
  journal = {arXiv preprint arXiv:2310.02238},
  year    = {2023}
}

@inproceedings{maini2024,
  title     = {{TOFU}: A Task of Fictitious Unlearning for {LLMs}},
  author    = {Maini, Pratyush and Feng, Zhili and Schwarzschild, Avi and Lipton, Zachary C. and Kolter, J. Zico},
  booktitle = {COLM},
  year      = {2024},
  note      = {arXiv:2401.06121}
}

@inproceedings{laine2024,
  title     = {Me, Myself, and {AI}: The Situational Awareness Dataset ({SAD}) for {LLMs}},
  author    = {Laine, Rudolf and Chughtai, Bilal and Betley, Jan and Hariharan, Kaivalya and Scheurer, Jeremy and Balesni, Mikita and Hobbhahn, Marius and Meinke, Alexander and Evans, Owain},
  booktitle = {NeurIPS Datasets and Benchmarks Track},
  year      = {2024},
  note      = {arXiv:2407.04694}
}

@inproceedings{park2023,
  title   = {Generative Agents: Interactive Simulacra of Human Behavior},
  author  = {Park, Joon Sung and O'Brien, Joseph and Cai, Carrie Jun and Morris, Meredith Ringel and Liang, Percy and Bernstein, Michael S.},
  booktitle = {Proceedings of the 36th Annual ACM Symposium on User Interface Software and Technology},
  pages   = {1--22},
  year    = {2023},
  doi     = {10.1145/3586183.3606763}
}

@inproceedings{brand2024,
  title   = {Using {GPT} for Market Research},
  author  = {Brand, James and Israeli, Ayelet and Ngwe, Donald},
  booktitle = {Proceedings of the 25th ACM Conference on Economics and Computation},
  pages   = {613--613},
  year    = {2024},
  doi     = {10.1145/3670865.3673479}
}

@article{grossmann2023,
  title   = {{AI} and the transformation of social science research},
  author  = {Grossmann, Igor and Feinberg, Matthew and Parker, Dawn C. and Christakis, Nicholas A. and Tetlock, Philip E. and Cunningham, William A.},
  journal = {Science},
  volume  = {380},
  number  = {6650},
  pages   = {1108--1109},
  year    = {2023},
  doi     = {10.1126/science.adi1778}
}

@article{bisbee2024,
  title   = {Synthetic Replacements for Human Survey Data? The Perils of Large Language Models},
  author  = {Bisbee, James and Clinton, Joshua D. and Dorff, Cassy and Kenkel, Brenton and Larson, Jennifer M.},
  journal = {Political Analysis},
  volume  = {32},
  number  = {4},
  pages   = {401--416},
  year    = {2024},
  doi     = {10.1017/pan.2024.5}
}

@article{tjuatja2024,
  title   = {Do {LLMs} Exhibit Human-like Response Biases? A Case Study in Survey Design},
  author  = {Tjuatja, Lindia and Chen, Valerie and Wu, Tongshuang and Talwalkwar, Ameet and Neubig, Graham},
  journal = {Transactions of the Association for Computational Linguistics},
  volume  = {12},
  pages   = {1011--1026},
  year    = {2024},
  doi     = {10.1162/tacl_a_00685}
}

@article{argyle2023,
  title   = {Out of One, Many: Using Language Models to Simulate Human Samples},
  author  = {Argyle, Lisa P. and Busby, Ethan C. and Fulda, Nancy and Gubler, Joshua R. and Rytting, Christopher and Wingate, David},
  journal = {Political Analysis},
  volume  = {31},
  number  = {3},
  pages   = {337--351},
  year    = {2023},
  doi     = {10.1017/pan.2023.2}
}

@article{bail2024,
  title   = {Can Generative {AI} improve social science?},
  author  = {Bail, Christopher A.},
  journal = {Proceedings of the National Academy of Sciences},
  volume  = {121},
  number  = {21},
  year    = {2024},
  doi     = {10.1073/pnas.2314021121}
}

@article{paulhus2003,
  title   = {The over-claiming technique: Measuring self-enhancement independent of ability},
  author  = {Paulhus, Delroy L. and Harms, P. D. and Bruce, M. Nadine and Lysy, Daria C.},
  journal = {Journal of Personality and Social Psychology},
  volume  = {84},
  number  = {4},
  pages   = {890--904},
  year    = {2003},
  doi     = {10.1037/0022-3514.84.4.890}
}

@article{atir2015,
  title   = {When Knowledge Knows No Bounds: Self-Perceived Expertise Predicts Claims of Impossible Knowledge},
  author  = {Atir, Stav and Rosenzweig, Emily and Dunning, David},
  journal = {Psychological Science},
  volume  = {26},
  number  = {8},
  pages   = {1295--1303},
  year    = {2015},
  doi     = {10.1177/0956797615588195}
}

@article{jung2026,
  title={PhantomBench: Benchmarking the Non-existential Threat of Language Models},
  author={Jung, Haeji and Gonen, Hila},
  journal={arXiv preprint arXiv:2606.11105},
  year={2026}
}

@inproceedings{zahraei2026,
  title={Prior Beliefs Prejudice {LLM}-as-Judge: Evidence from Persuasion Evaluation},
  author={Zahraei, Pardis Sadat and Wang, Xiaoning and Bozdag, Nimet Beyza and Tur, Gokhan and Hakkani-T{\"u}r, Dilek},
  booktitle={Findings of the Association for Computational Linguistics: ACL 2026},
  year={2026}
}

@article{lutz2025,
  title={The Prompt Makes the Person(a): A Systematic Evaluation of Sociodemographic Persona Prompting for Large Language Models},
  author={Lutz, Marlene and Sen, Indira and Ahnert, Georg and Rogers, Elisa and Strohmaier, Markus},
  journal={arXiv preprint arXiv:2507.16076},
  year={2025}
}

@inproceedings{ahn2024timechara,
  title     = {TimeChara: Evaluating Point-in-Time Character Hallucination of Role-Playing Large Language Models},
  author    = {Ahn, Jaewoo and Lee, Taehyun and Lim, Junyoung and Kim, Jin-Hwa and Yun, Sangdoo and Lee, Hwaran and Kim, Gunhee},
  booktitle = {Findings of ACL},
  year      = {2024},
  note      = {pages 3291--3325}
}

@inproceedings{sadeq2024rolefact,
  title     = {Mitigating Hallucination in Fictional Character Role-Play},
  author    = {Sadeq, Nafis and Xie, Zhouhang and Kang, Byungkyu and Lamba, Prarit and Gao, Xiang and McAuley, Julian},
  booktitle = {Findings of EMNLP},
  year      = {2024},
  note      = {pages 14467--14479}
}

@article{lin2025fallacies,
  title     = {Six Fallacies in Substituting Large Language Models for Human Participants},
  author    = {Lin, Zhicheng},
  journal   = {Advances in Methods and Practices in Psychological Science},
  year      = {2025},
  note      = {arXiv:2402.04470}
}

@inproceedings{seshadri2026lost,
  title     = {Lost in Simulation: {LLM}-Simulated Users are Unreliable Proxies for Human Users in Agentic Evaluations},
  author    = {Seshadri, Preethi and Cahyawijaya, Samuel and Odumakinde, Ayomide and Singh, Sameer and Goldfarb-Tarrant, Seraphina},
  booktitle = {ACL},
  year      = {2026},
  note      = {arXiv:2601.17087}
}

@inproceedings{yao2025taubench,
  title     = {$\tau$-Bench: A Benchmark for Tool-Agent-User Interaction in Real-World Domains},
  author    = {Yao, Shunyu and Shinn, Noah and Razavi, Pedram and Narasimhan, Karthik},
  booktitle = {International Conference on Learning Representations},
  year      = {2025},
  url       = {https://openreview.net/forum?id=roNSXZpUDN}
}

@inproceedings{bhat2025know,
  title     = {Know Thyself: Validating Knowledge Awareness of {LLM}-based Persona Agents},
  author    = {Bhat, Savita and Shukla, Ishaan and Karande, Shirish},
  booktitle = {Proceedings of the 5th Workshop on Trustworthy NLP (TrustNLP)},
  year      = {2025}
}

@inproceedings{gupta2024bias,
  title     = {Bias Runs Deep: Implicit Reasoning Biases in Persona-Assigned {LLM}s},
  author    = {Gupta, Shashank and Shrivastava, Vaishnavi and Deshpande, Ameet and Kalyan, Ashwin and Clark, Peter and Sabharwal, Ashish and Khot, Tushar},
  booktitle = {ICLR},
  year      = {2024}
}

\appendix
\section{Design of the Rating Experiment}

\paragraph{Items.} The battery holds 404 items: 178 valenced real items, their 178 screened
twins, and 48 filler items, which appear only in the two gate arms and enter
no estimate. 

\paragraph{Estimator algebra.}
Taking IPC separately on favourable- and unfavourable-content items,
\begin{equation}
d_{\mathrm{pos}} = \alpha f^{+} + \beta s^{+},
d_{\mathrm{neg}} = \alpha f^{-} + \beta s^{-}.
\end{equation}
Under additivity, coefficients ($\alpha$, $\beta$) common to both valence
groups, symmetric familiarity ($f^{+} = f^{-} = f$), and content of equal
magnitude and opposite sign ($s^{+} = -s^{-} = s$), the observed components
defined in \S\ref{sec:method} recover the two channels:
$F = (d_{\mathrm{pos}} + d_{\mathrm{neg}})/2 = \alpha f$ and
$S = (d_{\mathrm{pos}} - d_{\mathrm{neg}})/2 = \beta s$. Scale nonlinearity, asymmetric content
magnitude, or a familiarity--content interaction breaks the separation, so
the channel magnitudes are conditional on these assumptions. What does not
depend on them is the sign of $d_{\mathrm{neg}}$ itself, which is a direct
contrast rather than a derived quantity.

\paragraph{Persona conditions.}
Each of the six domains has three low-knowledge profiles (A1--A3) describing
distinct occupations, education levels, and interests, for 18 domain-specific
low-knowledge profiles in total. Each domain also has one expert profile (B;
six total) and a no-persona condition (C). All profiles are researcher-authored
and correspond to no real individuals.

\paragraph{Conditions.}
The gate-before condition asks the knowledge gate before the rating; the
gate-after condition asks it afterwards; the explicit condition writes
item-level ignorance into the profile.

\paragraph{Scale and counterbalancing.}
Seven-point scales presented in both directions, with reversed-direction
responses mapped back before analysis. The two paraphrase axes are not fully
crossed: one of the four wording cells is empty, so paraphrase enters as a
nuisance factor rather than an estimable one.

\paragraph{Inference.}
Confidence intervals are bootstraps over item clusters with 10,000 resamples,
because items repeat across trials while trials within an item are not independent.

\paragraph{Expert-condition rating ranges.}

Under the expert profile, per-endpoint mean ratings run $5.312$ to $5.714$
on favourable items and $2.031$ to $2.494$ on unfavourable ones, a
separation of at least $2.8$ rating points across these endpoint ranges.

\paragraph{Screening accounting.}

The stimulus bank generated 1,158 candidate twins. Screening discarded 697 because at
least one study model claimed to recognise the string in a pre-run recognition pass; 461
survived, and the run drew its 178 twins from that pool. Adversarial
realness scores informed construction, but the stored selected twins do not
uniformly minimize those scores among the recorded surviving candidates.

\paragraph{Screening records and matching limits.}
\label{sec:screenaudit}
The stored construction metadata reports exact-phrase checks against Wikipedia,
Wikidata, and OpenAlex on 12 August 2026. These are finite attestation checks,
not an exhaustive proof of nonexistence; the retained-twin recognition filter
uses the study endpoints themselves. Whole-bank metadata reports character-gram
AUROC values of $0.471$--$0.620$ across domains and token-count AUROC values of
$0.602$ for the OpenAI tokenizer proxy and $0.646$ for Gemini, both exceeding
the stated token-count target of $<0.60$. These bank-level summaries were not
recomputed for the 178 analysed pairs and do not establish equivalent matching
quality in that subset. Residual lexical differences remain a limitation.

\paragraph{Valence coding.}

Each real item carries a documented fact that makes its record favourable or
unfavourable, together with the reference page it was read from (Wikipedia or Wikidata
plus regulatory or news sources). Items whose documented record was mixed or contested
were not admitted to either tier. The coding is single-coder.

\paragraph{Trial counts and call parameters.}
Repeats are 2 question paraphrases $\times$ 3 draws $= 6$ per cell at
temperature 1.0, doubled for a small set of twins collected twice and
averaged before pairing (max tokens 16 for ratings, 200 for free-form; tools and
grounding disabled; one resample on parse failure); the model strings are the literal API
identifiers reported throughout.

\begin{table}[ht]
\caption{Rating-experiment design accounting. The two gate-order conditions were run across
five persona-condition slots per domain with unequal trial counts per slot;
the explicit condition used the three low-knowledge slots at equal size.}
\centering\small
\begin{tabular}{lrrr}
\toprule
Condition & Slots/domain & Trials/slot$^{*}$ & Trials \\
\midrule
gate, then rating & 5 & 9,660--12,360 & 56,400 \\
rating, then gate & 5 & 9,660--12,360 & 56,400 \\
item-level ignorance & 3 & 10,920 & 32,760 \\
\midrule
Total & & & 145,560 \\
\bottomrule
\end{tabular}
\par\smallskip{\footnotesize $^{*}$Aggregated over all domains and endpoints.}
\end{table}

\paragraph{Real--twin pairing accounting.} The analysed battery holds 178 valenced real items, 88 favourable-seeded and 90
unfavourable-seeded, each with one screened twin; no twin is reused across seeds.
Fourteen further real items had every twin candidate discarded in screening (8
favourable, 6 unfavourable). Since the estimator is defined only on a real--twin pair,
those items were dropped before rating and do not appear in the
analysis dataset. 

\paragraph{Parse failures.}
A response is parse-ok when it yields a single integer on the 1--7 scale; anything else
carries a parse-failure code and is excluded, never imputed. Of 145,560 trials, 453 (0.31\%) failed to parse: 266 with the gate first, 174 with the gate
last and 13 under the explicit sentence, so the primary condition has the
least missingness of the three.

\paragraph{Wording and scale-direction subgroups.}
Within the same-arm explicit-ignorance slice, $d_{\mathrm{pos}}$ is $+0.341$
under one instruction--question wording bundle and $+0.307$ under the other
(the two paraphrase axes were assigned jointly in this arm, so they alias and
cannot be separated), and $+0.298$ / $+0.358$ under normal / reversed scale
direction (88 seed pairs each). 

\section{Explicit-Exclusion Probe Extension}

The probe covers 24 real items and their twins on two endpoints (gpt-5.6-sol,
gemini-3.5-flash), each under two arms: the low-knowledge profile alone, and the same
profile plus the in-context sentence stating the persona has never heard the term. With
two profiles per domain (12 domain-specific profiles total) and 2 repeats this
gives 768 calls; the explicit-arm real cell holds 96
responses per endpoint. Responses were classified by rule (leading negation or a
no-knowledge marker within the first 80 characters) and checked by a second pass.

\end{document}